\documentclass[lettersize,journal]{IEEEtran}
\usepackage{amsmath,amsfonts}
\usepackage{algorithmic}
\usepackage{algorithm}
\usepackage{array}
\usepackage[caption=false,font=normalsize,labelfont=sf,textfont=sf]{subfig}
\usepackage{textcomp}
\usepackage{stfloats}
\usepackage{url}
\usepackage{verbatim}
\usepackage{graphicx}
\usepackage{cite}
\usepackage[table,xcdraw]{xcolor}
\usepackage{amsmath,amsfonts}

\usepackage{newfloat}
\usepackage{listings}
\usepackage{xcolor} 
\usepackage{booktabs} 
\usepackage{multirow}
\usepackage{graphicx}
\usepackage{amsmath}
\usepackage{amssymb}
\usepackage{booktabs}
\usepackage{array}  
\usepackage{enumitem}
\usepackage{amsmath}
\usepackage{amssymb}

\newcommand{\bp}{\boldsymbol{p}}
\newcommand{\bo}{\boldsymbol{o}}

\newcommand{\bz}{\boldsymbol{z}}

{}
\begin{document}

\title{Point Cloud Diffusion with Global and Local Reconstruction for Instance-Level 3D Anomaly Detection}

\author{Linchun~Wu,
	Qin~Zou,~\IEEEmembership{Senior Member,~IEEE},
	Jiwen Lu,~\IEEEmembership{Fellow,~IEEE},
		Qingquan~Li
\thanks{{L.~Wu and Q.~Zou are with the School of Computer Science, Wuhan University, Wuhan 430072, China (E-mails: \{linchun.wu, qzou\}@whu.edu.cn).}}
\thanks{J.~Lu is with  Department of Automation, Tsinghua
	University, Beijing 100084, China (e-mail: lujiwen@tsinghua.edu.cn).}
\thanks{Q.~Li is with the Guangdong Artificial Intelligence and Digital Economy Laboratory (SZ), Shenzhen 518060, China (e-mail: liqq@szu.edu.cn).}
}

\markboth{Journal of \LaTeX\ Class Files,~Vol.~xx, No.~xx,  April~2026}%
{Shell \MakeLowercase{\textit{et al.}}: A Sample Article Using IEEEtran.cls for IEEE Journals}


\maketitle

\begin{abstract}
	3D anomaly detection in point clouds is critical for high-precision industrial manufacturing. Reconstruction-based methods have laid a strong foundation by detecting 3D anomalies through comparisons between defective inputs and their reconstructed normal counterparts. However, existing methods still suffer from two  challenges: 1) the foreground weak defective regions such as scratches are hard to reconstruct and detect, where the anomaly deviations in normalized point clouds can be as small as $10^{-3}$; 2) the background non-defective regions are prone to get positional bias in reconstruction, which leads to false positives. To address these challenges, we propose \textbf{PCDiff}, a point cloud diffusion framework for instance-level 3D anomaly generation and detection. In the generation phase, an instance-level multi-modal attention is embedded into the generation framework, where anomalies are  conditioned with texture gradient, image patch, text and mask. The instance-level condition enables the high-quality generation of weak-defective anomalies. In the detection phase, a joint local-global reconstruction algorithm is introduced to ensure local anomaly restoration and global geometric consistency, which preserves background normal structure while restoring the foreground defect. Extensive experiments demonstrate that the proposed PCDiff significantly outperforms state-of-the-art methods in both 3D anomaly generation fidelity and reconstruction quality, leading to substantial improvements in anomaly detection accuracy.
\end{abstract}


\section{Introduction}
3D data, such as point clouds and depth maps acquired by structured-light or LiDAR sensors, have become increasingly prevalent in multimedia systems—not only for entertainment and communication but also for industrial multimedia applications like automated visual inspection~\cite{11396975, liu2024real3d, 10922728} and smart manufacturing. In this context, \textbf{3D anomaly detection} is vital for maintaining structural integrity, identifying functional defects—such as volumetric stability of energy cells~\cite{tang2024cloud} to the intricate topological continuity of high-density electronics~\cite{wang2024anomaly}. While legacy 2D paradigms~\cite{11360496, 10443076} are limited by the loss of spatial dimensionality, the shift toward point cloud-based modalities represents a paradigm leap. By providing a high-fidelity digital depiction of physical surfaces, point clouds enable the robust synthesis and identification of subtle structural irregularities that deviate from nominal patterns~\cite{ liang2025look}.

\begin{figure}[!t]
	\centering
	\includegraphics[width=1\linewidth]{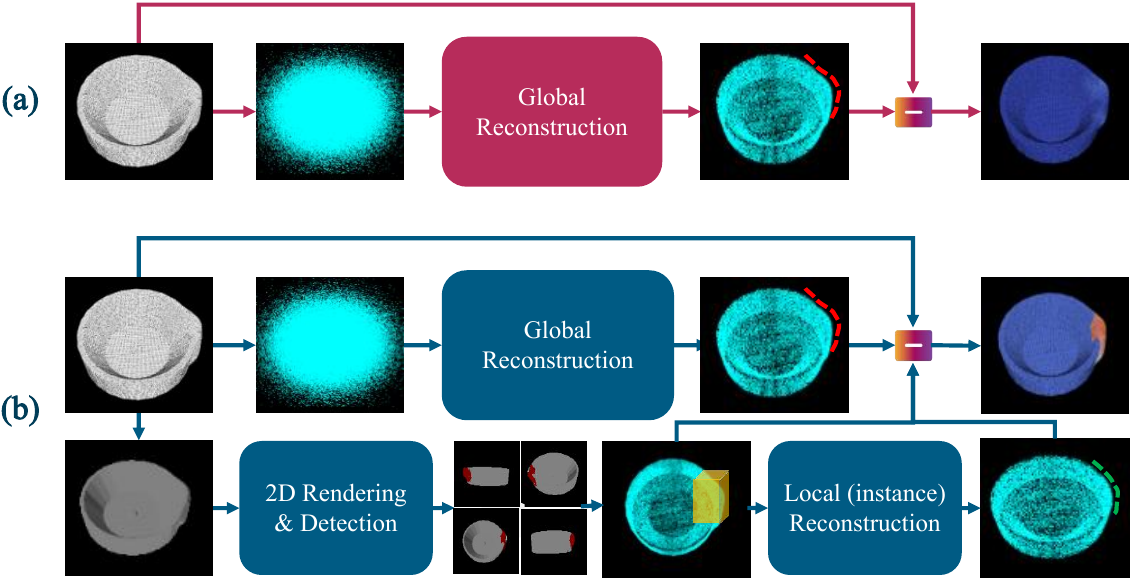}
	\caption{Two different pipelines for 3D anomaly detection. (a) The conventional pipeline that performs a global reconstruction across all points, lacking spatial adaptivity to weak anomaly regions. (b) The proposed pipeline that integrates global and local reconstruction. The joint reconstruction preserves background normal structure while restoring the foreground defects. 
	}
	\label{fig:idea}
\end{figure}
3D multimedia processing has seen significant growth, driven by advances in geometric representation learning~\cite{10970054, li2024towards, rudolph2023asymmetric, zhou2024r3d, liang2025examining}. A core challenge lies in the fine-grained decoupling of benign manufacturing variance from functional defects. Since real-world items exhibit inherent geometric noise that does not impair functionality, distinguishing these from true 3D anomalies---such as protrusions or pinholes---requires sophisticated generative patterns. Reconstruction-based paradigms address this by leveraging the statistical distribution of normal samples to identify defects through the residual discrepancy between input signals and their ``normal'' reconstructions~\cite{li2024towards, rudolph2023asymmetric}.

The success of diffusion models in conditional 3D generation~\cite{zhou2024r3d,jin2024dualanodiff} has inspired their application in anomaly detection. R3D-AD~\cite{zhou2024r3d} employs progressive denoising for point cloud reconstruction, while PO3AD~\cite{ye2024po3ad} improves representation learning by predicting deviations without explicit noise injection. Despite these gains, many diffusion-based frameworks rely on global reconstruction over the entire point set (Fig.~\ref{fig:idea}). This global approach lacks spatial adaptivity, particularly for subtle or weak anomalies, leading to two critical challenges that limit detection performance and practical applicability.

\begin{figure}[!t]
	\centering
	\includegraphics[width=1\linewidth]{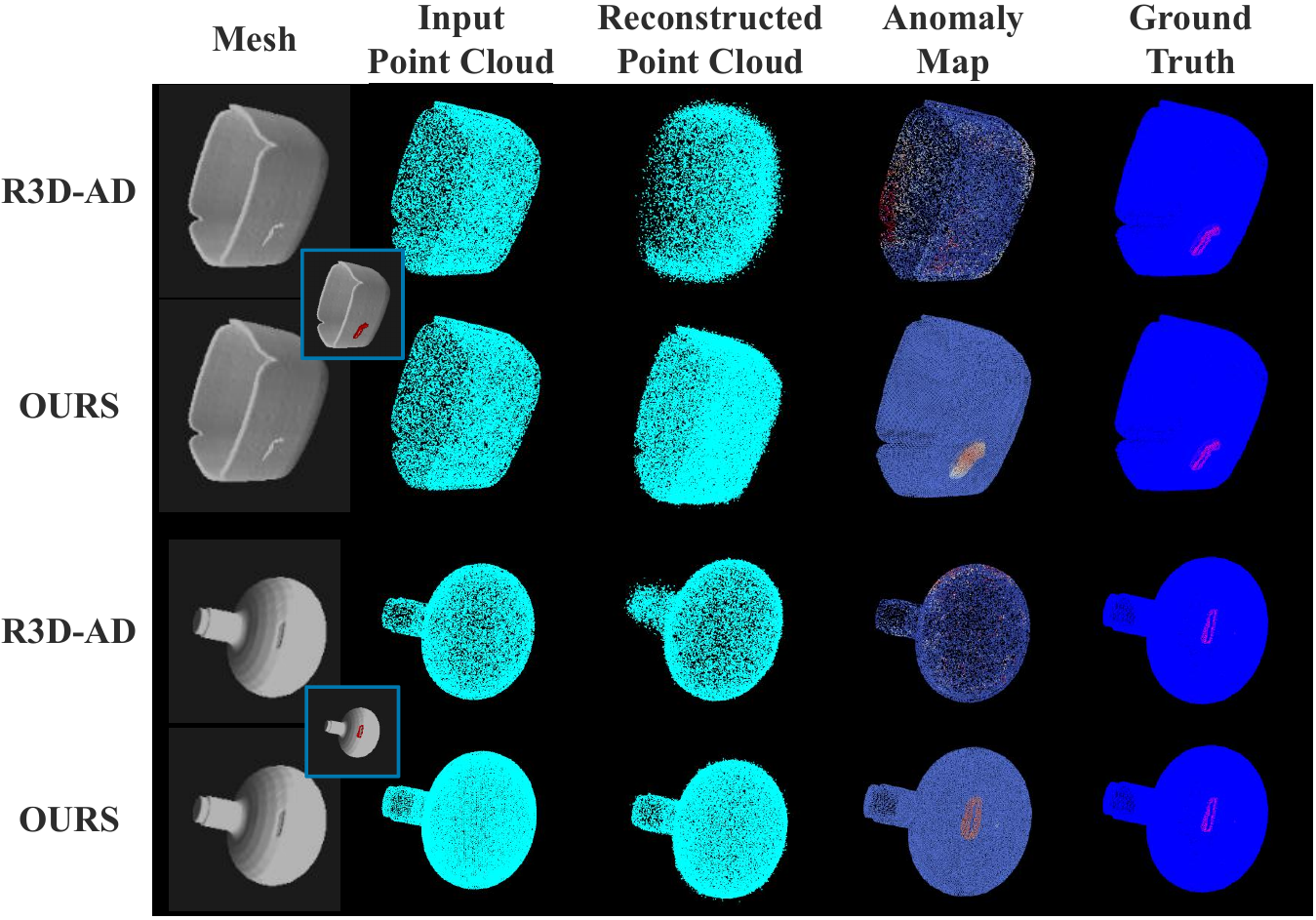}
	\caption{Comparison of reconstruction results obtained by R3D-AD and PCDiff. The reconstructions produced by R3D-AD exhibit noticeable coarseness, with originally sharp edges appearing rounded and point positions significantly biased.  The proposed PCDiff preserves the geometric integrity of non-defective regions while reconstructing anomalous areas, resulting in more accurate anomaly maps.
	}\label{fig:show_result}
\end{figure}

First, weak foreground regions like fine scratches are difficult to reconstruct and detect. Current 3D pseudo-anomaly generators often rely on high-amplitude geometric deformations with smooth boundaries~\cite{zhou2024r3d}, failing to model the linear or texturally complex defects found in real-world scenarios. Consequently, models struggle to identify small-scale anomalies (Fig.~\ref{fig:show_result}). 

Second, background regions often suffer from positional bias during reconstruction, causing false positives. Standard strategies favor global structural fidelity over local texture consistency~\cite{zhou2024r3d, li2024towards}, leading to over-smoothed edges and distorted point distributions in normal areas. Addressing these issues requires an anomaly-aware reconstruction that prioritizes restoring defective regions while preserving normal geometry.

To tackle these challenges, we propose \textbf{PCDiff}, a unified diffusion framework for 3D anomaly generation and detection. For generation, PCDiff adopts a multi-view image-guided pipeline that utilizes an instance-level multi-modal attention mechanism to incorporate texture gradients, image patches, and spatial masks. For reconstruction, PCDiff leverages 2D anomaly mask estimation derived from rendered shadow differences to localize subtle defects. These masks guide a local-global joint reconstruction module, where a local branch focuses on anomaly restoration and a global branch ensures holistic geometry recovery.

Our main contributions are summarized as follows:

\begin{itemize}[leftmargin=*]
	\item We propose \textbf{PCDiff}, a novel framework leveraging 2D visual priors and multi-modal conditioning to synthesize fine-grained anomalies and achieve robust instance-level detection.
	\item We introduce a geometry bank-guided synthesis method combined with gradient-based texture representations to generate diverse, subtle anomalies representing real-world defects.
	\item We design a local-global joint reconstruction framework that integrates image-derived priors for focused anomaly restoration and global geometry recovery, enhancing both accuracy and spatial consistency.
\end{itemize}

\noindent The rest of the paper is organized as follows. Section~\ref{sec:related_work} reviews relevant work. Section~\ref{sec:generation} details the proposed \textbf{PCDiff} generation framework. Section~\ref{sec:detection} illustrate the proposed local-global joint detection module. Section~\ref{sec:experiments} presents experiments and results. Section~\ref{sec:conclusion} concludes the work. 

\section{Related Works}
\label{sec:related_work}

\subsection{3D Generation}

Modern 3D generative models utilize diverse representations, including point clouds~\cite{luo2024scalable, 10907786, 10944293, liu2026hifi3d}, triplanes~\cite{cao2024difftf++}, NeRF~\cite{lee2024disr}, and Gaussian splatting~\cite{tang2023dreamgaussian}. Current frameworks typically employ diffusion~\cite{10934729} either directly on 3D geometry~\cite{zhang2024clay} or within a VAE-encoded latent space~\cite{xiang2025structured}. While two-stage methods~\cite{zhang2024clay} produce artistic textures via per-face generation, they incur high computational costs. Conversely, one-stage methods~\cite{xiang2025structured} enhance efficiency by projecting image features into the latent space for computationally feasible synthesis.

Optimization-based methods~\cite{melas2023realfusion, metzer2023latent} leverage 2D diffusion priors for high-quality multi-view generation. By optimizing 3D distributions through rendering losses, these approaches achieve impressive results but often suffer from prolonged optimization, ``multi-face'' artifacts, and limited diversity~\cite{liu2024one}. To improve quality, multi-view information has been integrated across various modalities, including meshes~\cite{zhang2024clay}, triplanes~\cite{cao2024difftf++}, and Signed Distance Fields~\cite{liu2024one}. Clay~\cite{zhang2024clay}, for instance, utilizes a compressed latent-space voxel cloud to enable multi-modal control over text, images, and point clouds.

Despite these advancements, existing models struggle with the fine-grained, localized synthesis required for anomaly generation. Achieving instance-level control over category, region, and texture remains a critical challenge for realistic defect synthesis.

\subsection{3D Anomaly Detection}

Current 3D anomaly detection methods generally follow either reconstruction-based or feature embedding-based paradigms. {Reconstruction-based methods}~\cite{li2024towards, wang2023multimodal} operate on the assumption that defective regions can be restored to a normal state. These frameworks typically mask portions of the input and measure the residual discrepancy between the original signal and its normal reconstruction. Building on this, AST~\cite{rudolph2023asymmetric} utilizes an asymmetric teacher-student network, hypothesizing that anomalous samples---absent from the training set---will yield larger output distances. However, the reliability of these models remains constrained by the lack of explicit anomaly knowledge during training.

To improve robustness, {feature embedding methods}~\cite{liu2025duinnet, liu2024real3d, wang2023multimodal} leverage reference normal samples. Reg3D-AD~\cite{liu2024real3d} employs memory banks to retrieve similar normal features, while Shape-guided~\cite{chu2023shape} integrates multi-modal memory banks from both images and point clouds. These approaches enhance reconstruction stability and interpretability by reducing the domain gap between training and inference. For multi-modal fusion, BTF~\cite{horwitz2023back} identifies optimal feature representations for specific defect types, strategically combining 2D and 3D detection to significantly boost accuracy.

\noindent \textbf{Remark.} The absence of anomaly knowledge during training remains a critical bottleneck. Our approach directly addresses this by synthesizing artificial anomalies to bridge the information gap, ensuring more reliable model outputs and enhanced interpretability.
\begin{figure*}[!t]
	\centering
	
	\includegraphics[width=1\linewidth]{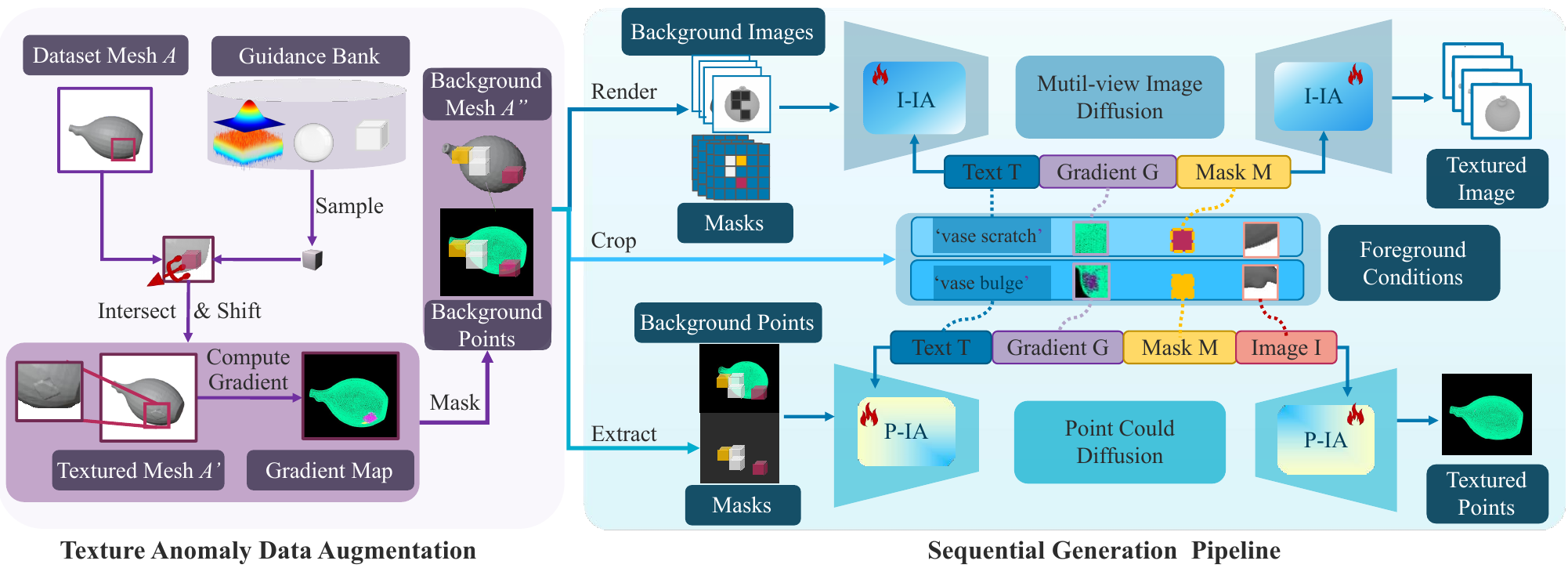}
	
	\caption{An overview of the proposed PCDiff.  The left part illustrates texture augmentation process. The input datasets mesh $A$ is augmented by inserting a random mesh from the guidance bank. A textured mesh $A'$ is created by shifting the intersecting vertices. After gradient representation, the instance mask is applied. Then we render multi-view image for image generation and extract point cloud for 3D point diffusion. Instances are cropped based on the mask for both modalities. The right part depicts the sequential generation. Multi-view image generation takes masked images and the mask as reference to produce textured multi-view images. Instance conditioning in this stage includes text {\it T}, colored gradient images {\it G}, and the mask {\it M}, which are used in image-instance attention, I-IA. In the point cloud generation, masked point clouds and point masks serve as input, generating textured anomaly point clouds as output. In point-instance attention P-IA, the cropped image {\it I} is added as a condition.}
	
	\label{fig:framework}
\end{figure*}
\section{Textured Anomaly Generation}
\label{sec:generation}

Our framework facilitates 3D anomaly detection by integrating multi-view and point cloud generation to provide diverse data augmentation. This pipeline consists of two primary components: \textbf{Texture Augmentation}, which synthesizes subtle texture-level defects, and \textbf{Sequential Generation}, which ensures high-fidelity cross-modal alignment. To ensure comprehensive coverage, we also incorporate an off-the-shelf method~\cite{liang2025examining} for detecting prominent structural anomalies.

\noindent\textbf{Texture Augmentation.} 
As depicted in Fig.~\ref{fig:framework}, this module synthesizes controllable anomalies by perturbing mesh surfaces. We simulate geometric deformations by intersecting a guidance geometry with a non-defective mesh region and perturbing vertices along their normals. A computed gradient map provides conditioning for texture-aware diffusion. The resulting mesh is partitioned into three instances: \texttt{anomaly} (e.g., bulge), \texttt{texture-augmented} (e.g., scratch), and \texttt{normal}. These meshes and gradient maps are rendered into multi-view images for 2D diffusion, while corresponding point clouds are extracted for 3D generation. All inputs are cropped to ensure consistent instance-level conditioning.

\noindent\textbf{Sequential Generation.} 
To achieve high-fidelity reconstruction, 3D point cloud diffusion utilizes multi-view images as spatial priors. The view with maximum pixel coverage of the target instance is selected as the primary condition for the 3D branch. We introduce a shared Instance Attention (IA) module across both pipelines to enable controllable generation of appearance, category, and spatial location. In the 2D pipeline, masked multi-view images $I_r$ are conditioned on a mask $M$, text prompt $c_\mathcal{T}$, and gradient map $c_G$. For 3D diffusion, masked point features $P_r$ are augmented with $M$, $c_\mathcal{T}$, $c_G$, and a cross-modal image patch $c_I$.

\subsection{Texture Augmentation}
We synthesize textures by perturbing 3D mesh vertices, categorized into irregular and structured patterns. \textbf{Irregular textures} are generated via isotropic Gaussian noise. \textbf{Structured textures} utilize a geometry-guided strategy leveraging a guidance bank of primitives (cubes for linear edges, spheres for curved contours). By applying per-instance scaling, rotation, and spatial overlay, we generate diverse anomaly patterns.

\subsubsection{Geometric Guidance and Scaling}
We sample $n \in \{1, 2, 3\}$ primitives $G = \{g_1, \dots, g_n\}$. To maintain statistical consistency, each $g_i$ is scaled by $s$, constrained within $[\max(\bar{r} - r, 0), \min(\bar{r}, r)]$. Here, $r$ is the local anomaly size of the input, and $\bar{r}$ is the dataset-level mean anomaly scale:
\begin{equation}
	\bar{r} = \frac{1}{3N} \sum_{j=1}^{N} \sum_{d \in \{x, y, z\}} r_{j}^d,
\end{equation}
where $r_{j}^d$ represents the bounding box half-length along axis $d$.

\subsubsection{Spatial Transformation and Perturbation}
Each primitive is transformed via a random rotation matrix $R_i$ and translated to an anchor vertex $v$ sampled from non-defective regions:
\begin{equation}
	g''_i = (R_i \times (g_i \cdot s)) + v.
\end{equation}

The intersection of $g''_i$ and the mesh defines the perturbation region. For each vertex $p_i$ within this volume, we apply a normal displacement:
\begin{equation}
	\hat{p}_i = p_i + \eta \cdot \mathbf{n}_i,
\end{equation}
where $\eta \sim \mathcal{U}(-0.05, 0.05)$ represents the displacement magnitude, with the bounds derived from empirical dataset observations.

\subsection{Transformation-based Texture Representation}
To facilitate both 2D image generation and 3D point cloud diffusion, we propose a \textbf{point-based, transformation-aware texture representation}. This descriptor encodes the transition from smooth surfaces to augmented anomalies by leveraging the vertex displacement field. Specifically, for each vertex $i$, the displacement vector is defined as $\Delta p_i = \hat{p}_i - p_i$.

To capture local geometric variations, we compute a localized gradient field. The gradient $\nabla_{g_i}$ at vertex $i$ is normalized by the average distance to its $k$ nearest neighbors within a local neighborhood:
\begin{equation}
	\nabla_{g_i} = \frac{ | \hat{p}_i - p_i | }{ \frac{1}{k} \sum_{j=1}^{k} \| p_i - p_j \| },
\end{equation}
where $p_j$ denotes the $j$-th neighboring vertex. This formulation explicitly encodes the directional shifts and relative magnitudes of the surface perturbations. By capturing these transformation dynamics, the representation provides a discriminative feature set that enhances cross-modal adaptability for generative tasks.

\subsection{Multi-modal Instance Attention}

This module is designed to generate anomalies with precise control over localization, category, and texture. For both 2D and 3D generation, we design a
 unified structure for modality fusion and instance allocation.  

Specifically, each anomaly instance is defined by a text prompt $c_\mathcal{T}$, a gradient map $c_G$ (an image in 2D generation or points in 3D generation), a regional mask $m$, and optionally a cropped image $c_I$.  
The modality fusion process is formulated as:
\begin{equation}
	c = \text{MLP}\left([\tau_\theta(c_\mathcal{T}), \gamma_\theta(c_I)]\right),
\end{equation}
where $\tau_\theta$ and $\gamma_\theta$ denote the CLIP text and image encoders, respectively, and \text{MLP} serves as the fusion layer.

To enable fine-grained instance allocation, we adopt a mask-guided attention mechanism that selectively integrates features from different modalities. The formulation is:  
\begin{equation}
	\hat{Z} = \Bigg(\sum_{j=0}^{N} \text{Attn}\big(Q^f_j, K^c_j, V^c_j, M_j\big) + G_j \Bigg) + Z \cdot \overline{M},
\end{equation}
where $j$ indexes the anomaly instance and $N$ is the total number of instances. $Q^f_j$ denotes the residual noise feature, while $K^c_j$ and $V^c_j$ are the key and value features derived from the fused condition $c$. The gradient conditioning is given by $G_j = \rho_\theta(c_G)$, with $\rho_\theta$ representing the gradient encoder. $M_j$ is the instance attention mask, where non-instance regions are set to $-\infty$ and instance regions to $1$. $Z$ is the normalized input, and $\overline{M}$ is the complementary mask, where non-instance regions are $1$ and instance regions are $0$.  

This design ensures that each anomaly instance receives condition-specific attention while maintaining separation between background and instance regions. The combination of modality fusion and mask-guided allocation enables precise, controllable, and semantically aligned anomaly generation across both 2D and 3D domains.

\begin{figure*}[!t]
	\centering
	
	\includegraphics[width=0.9\linewidth]{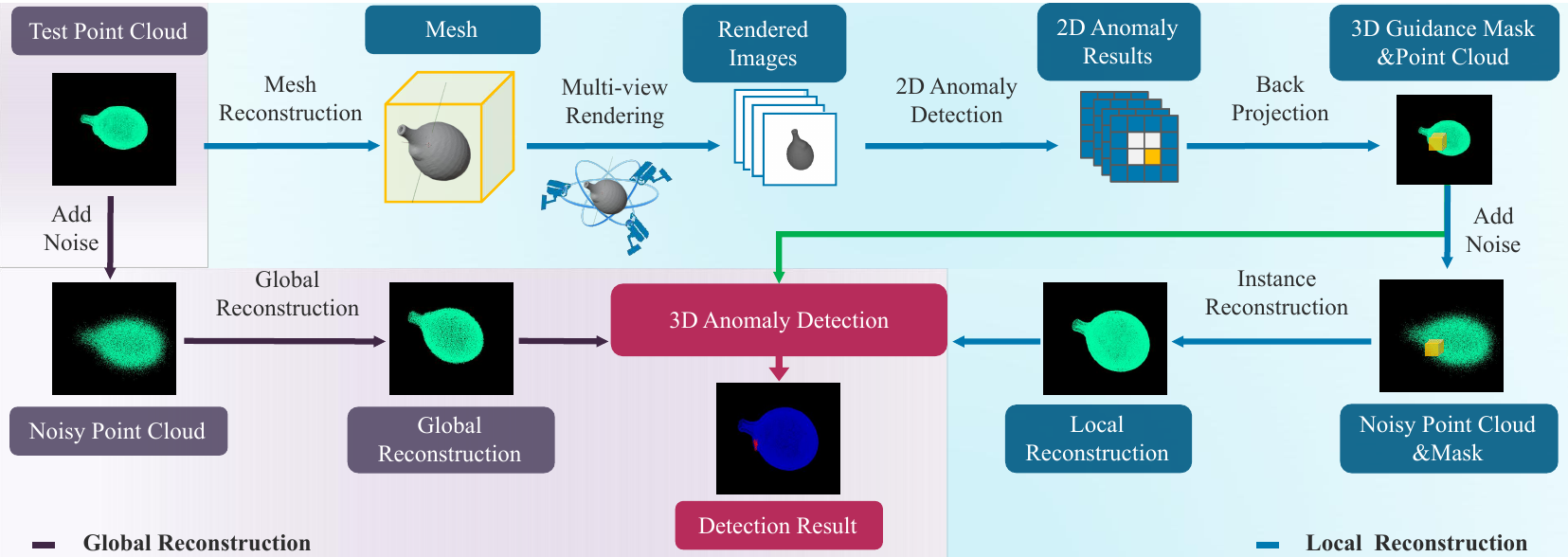}
	
	\caption{The local-global anomaly detection framework. The local path reconstructs the test point cloud into a mesh, renders multi-view images, and estimates anomalies via comparison of rendered image and reconstructed image. The estimate image anomaly masks are then projected into 3D space as point masks, guiding instance-level reconstruction. The global path applies integral noise for unbiased geometry reconstruction. The local and global results are merged for final anomaly detection.
	}
	
	\label{fig:detection}
\end{figure*}

\subsection{Training Objectives}

For multi-view generation, the training objective can be formulated as:
\begin{equation}
	\begin{aligned}
		\mathcal{L}_{MDH} = E_{\varepsilon(x),\varepsilon({x_r}),\epsilon \sim N(0,\boldsymbol{I}),t}||\epsilon_t -\epsilon_{\theta({\bz(t)},t,m,\boldsymbol{\mathcal{T}}, G} ||^2,
	\end{aligned}
	\label{eq: diffusion_loss}
\end{equation}
where $x$ denotes the anomaly image, $x_r$ represents the masked image reference, $m$ represents the mask, ${\bz }$ is the noisy image feature pre-encoded with VQVAE $\varepsilon$, $G$ is the gradient map, $t$ is a random diffusion time step, and $\boldsymbol{\mathcal{T}}$ is text embedding.

For point cloud generation, we extract and quantize the point cloud from the augmented mesh with a 0.01 voxel size. The sample includes masked point cloud ${p_r}$ and sampled point clouds ${p}$, with the multi-view image $I$ and texture map $G$ as control. The training objective is as follows:
\begin{equation}
	\begin{aligned}
		\mathcal{L}_{PDH} = E_{\bp,{\boldsymbol{{p_r}}},\epsilon \sim N(0,\boldsymbol{I}),t}||\epsilon_t -\epsilon_{({\bp(t)},t,m,\boldsymbol{\mathcal{T}}, G,I)} ||^2.
	\end{aligned}
	\label{eq: diffusion_loss}
\end{equation}

\section{Local-Global Anomaly Detection} 
\label{sec:detection}

As illustrated in Figure~\ref{fig:detection}, we introduce a {local-global anomaly detection framework} designed to achieve anomaly-aware reconstruction and robust multi-modal detection. The local branch emphasizes instance-level reconstruction guided by 2D-predicted anomaly regions, whereas the global branch performs unbiased reconstruction of the complete point cloud. By jointly leveraging the local reconstruction point cloud, the global reconstruction point cloud, and the image-level anomaly masks, the final detection stage achieves comprehensive and reliable anomaly point identification.

\subsection{Local Reconstruction}
\label{sec:appendix_reconstruction}

\label{sec:rendering}
To achieve comprehensive surface coverage for anomaly localization, we define a structured multi-view acquisition setup. We employ $N=3$ circular camera trajectories at elevation angles of $45^\circ$, $0^\circ$, and $-45^\circ$. Along each trajectory, $K=6$ cameras are distributed at equal intervals ($60^\circ$ azimuthal steps), totaling 18 views per object. The rendering operation for a viewpoint $(i,j)$ is defined as:
\begin{equation}
	\mathcal{I}_{i,j} = \mathcal{R}(M, \mathcal{J}_{i,j}),
\end{equation}
where $\mathcal{J}_{i,j} \in SE(3)$ represents the camera pose matrix. This configuration ensures that even subtle anomalies located in occluded regions (e.g., the underside of a vase) are captured by at least two viewing frustums.

The back-projection operator $\Pi^{-1}$ maps pixel-level anomaly probabilities from the image plane back to the input point cloud $P$. For a point $p \in P$, the mapping is determined by:
\begin{equation}
	\Pi^{-1}(U, P, \mathcal{J}) = U(\pi(\mathcal{J} \cdot p)),
\end{equation}
where $\pi$ denotes the perspective projection function. Points falling outside the camera frustum or those occluded by the mesh geometry $M$ (determined via depth-buffer testing) are assigned a score of zero for that specific view. Averaging these scores across all $N \times K$ views effectively filters out 2D detection noise and highlights regions with high multi-view consensus.

Finally, the fused anomaly prior \( U_p \) guides an anomaly-aware reconstruction network \( \psi \), defined as:
\begin{equation}
	\hat{P}_{\text{l}} = \psi(P, U_p),
\end{equation}
where \( \hat{P}_{\text{l}} \) is the locally reconstructed point cloud emphasizing anomaly-prone regions. By steering the network to attend to structurally or texturally abnormal areas, the local branch ensures high-fidelity recovery of subtle anomalies.

\subsection{Global Reconstruction}

To achieve comprehensive 3D reconstruction, we incorporate a global reconstruction branch that operates on the complete input point cloud:
\begin{equation}
	\hat{P}_{g} = \psi(P),
\end{equation}
\noindent where \( \hat{P}_{g} \) denotes the globally reconstructed point cloud, and \( \psi \) represents a reconstruction network derived from the base generative framework, omitting multi-modal instance attention mechanisms. This global pathway captures holistic geometry without relying on anomaly priors, ensuring unbiased structural restoring.


\subsection{Training Objectives}

To accelerate reconstruction and preserve authentic input geometry, avoiding plausible but incorrect shape hallucinations, we introduce a one-step offset prediction objective. It learns to map anomalous coordinates to their normal counterparts. The training loss is:
\begin{equation}
	\mathcal{L}_{R} = \mathbb{E}_{(\bp), \bo} \left\| \bo - \bo_{\left( \bp,  m \right)} \right\|^2,
	\label{eq: diffusion_loss}
\end{equation}
\noindent where \( \bo \) is the pseudo ground-truth offset between noisy anomaly and clean normal coordinates, ${\bp }$ is the noisy point cloud, and \( m \) is an optional anomaly-aware mask. This encourages learning robust geometric corrections under defective inputs.

\subsection{Merging and Detection}

For comprehensive 3D anomaly detection, we fuse local reconstruction, global reconstruction, and multi-view image-derived anomaly priors. The local branch restores fine-grained structures within anomalies, while the global branch provides anomaly-agnostic structural consistency. Concurrently, the image-derived mask captures 2D texture cues. 

To integrate these sources, we adopt a selective merging strategy. The final anomaly score \( S_a \) is:
\begin{equation}
	S_a = \zeta_{\nu} \left( P,\; \hat{P}_l \cdot U_p + \phi(\hat{P}_l + \hat{P}_g) \cdot \overline{U_p},\; U_p \right),
\end{equation}
\noindent where \( \zeta_{\nu} \) is the detection network, \( \hat{P}_l \) and \( \hat{P}_g \) are local and global reconstructions, \( U_p \) is the 3D anomaly prior from image space, and \( \overline{U_p} = 1 - U_p \). The fusion layer \( \phi(\cdot) \) (two transformer blocks) adaptively merges reconstructions, ensuring accurate anomaly attribution by dynamically balancing local fidelity and global coherence.

\begin{figure*}[!t]
	\centering

		\includegraphics[width=\linewidth, height=0.4\textheight]{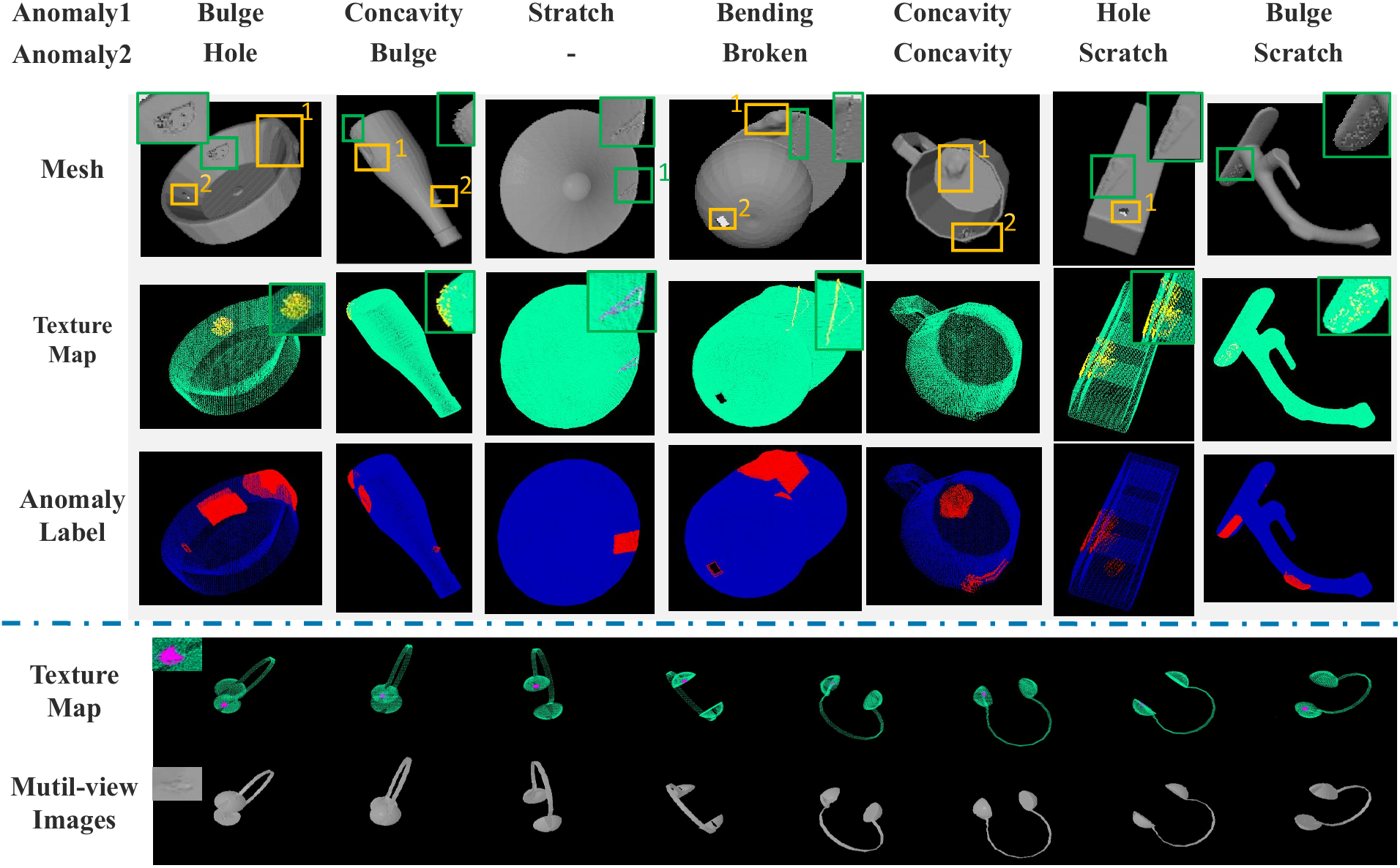}
	\caption{3D anomaly generation results obtained by the proposed PCDiff. Top: textured 3D anomaly samples. Bottom: the generated multi-view images.  }
	\label{fig:compare}
\end{figure*}
\section{Experiments}
\label{sec:experiments}
\subsection{Datasets}

\noindent\textbf{Anomaly-ShapeNet} Anomaly-ShapeNet~\cite{li2024towards} comprises a total of 1,600 samples which are distributed across 40 distinct categories. There are six kinds of anomalies, including bulge, concavity, bending, crack, hole, and broken.

\begin{table}[h!]
	\centering
	\renewcommand{\arraystretch}{0.8} 
	\setlength{\tabcolsep}{7pt} 
	\caption{Comparison of generation methods across F-Score, CLIP similarity, and categories on Anomaly-ShapeNet dataset.}
	\begin{tabular}{@{}l|m{1.8cm}m{1.8cm}m{1.7cm}@{}}
		\toprule
		\textbf{Method} & \textbf{F-Sco.(\%)} $\uparrow$ & \textbf{CLIP-Sim} $\uparrow$  & \textbf{Categories} $\uparrow$ \\ 
		\midrule 
		Gau-SP ~\cite{kerbl20233d}      & 38.2 & 54.6  & 6 \\ 
		CutPaste~\cite{li2021cutpaste} & 81.6 & 63.1  & 6 \\ 
		R3D-AD~\cite{zhou2024r3d}     & 89.7 & 77.3  & 3 \\ 
		Ours                     & 95.4 & 81.0  & 6 \\ 
		\bottomrule
	\end{tabular}
	\vspace{-5pt}
	\label{tab:generation}
\end{table}

\noindent\textbf{Real3D-AD} Real3D-AD~\cite{liu2024real3d} comprises a total of 1,254 samples that are distributed across 12 distinct categories. These categories include Airplane, Car, Candybar, Chicken, Diamond, Duck, Fish, Gemstone, Seahorse, Shell, Starfish, and Toffees. 

For both dataset, each training set for a category contains only four samples while the test set contains normal and various defect samples.

\begin{table*}[t]
	\centering
	\footnotesize
	\caption{O-AUROC / P-AUROC scores for anomaly detection across 15 categories of Anomaly-ShapeNet. Bold indicate the best results.}
	\label{tab:anoshape}
	\setlength{\tabcolsep}{3.2pt}
	\renewcommand{\arraystretch}{1.1}
	\begin{tabular}{l|cccccccccc}
		\toprule
		\textbf{Category} & \textbf{M3DM\cite{wang2023multimodal}} & \textbf{IMRNet\cite{li2024towards}} & \textbf{R3D-AD\cite{zhou2024r3d}} & \textbf{ISMP\cite{liang2025look}} & \textbf{PO3AD\cite{ye2024po3ad}} & \textbf{PASDF\cite{Zheng2025Bridging3D}} & \textbf{MC3AD\cite{cheng2025mc3d}} & \textbf{MC4AD\cite{liang2025examining}} & \multirow{2}{*}{\cellcolor{pink!20}\textbf{Ours}} \\
		
		\textbf{Source} & \textbf{CVPR'23} & \textbf{CVPR'24} & \textbf{ECCV'24} & \textbf{AAAI'24} & \textbf{CVPR'25} & \textbf{ICCV'25} & \textbf{IJCAI'25} & \textbf{NuerlIPS'25} & \\
		\midrule
		Feat. & PMAE & PMAE & Raw & Raw & Raw & Raw & PointMAE & Raw & \cellcolor{pink!20}Raw \\
		\midrule
		Ashtray    & 0.58/0.58 & 0.67/0.67 & 0.83/- & 0.91/0.60 & \textbf{1.00}/0.96 & \textbf{1.00}/0.92 & 0.96/- & \textbf{1.00}/0.90 & \cellcolor{pink!20}\textbf{1.0}/\textbf{0.98} \\
		Bag        & 0.54/0.64 & 0.66/0.67 & 0.72/- & 0.75/0.74 & 0.83/0.5  & \textbf{0.99}/\textbf{0.96} & 0.81/- & 0.98/\textbf{0.96} & \cellcolor{pink!20}0.93/0.90 \\
		Bottle     & 0.58/0.61 & 0.63/0.63 & 0.75/- & 0.78/0.67 & 0.92/0.88 & \textbf{1.00}/0.94 & 0.75/- & 0.97/0.91 & \cellcolor{pink!20}\textbf{1.00}/\textbf{0.95} \\
		Bowl       & 0.58/0.63 & 0.68/0.68 & 0.75/- & 0.78/0.72 & 0.88/0.94 & 0.97/0.90 & 0.86/- & 0.94/0.96 & \cellcolor{pink!20}\textbf{0.98}/\textbf{0.96} \\
		Bucket     & 0.41/0.70 & 0.68/0.68 & 0.72/- & 0.73/0.60 & 0.82/0.83 & 0.87/0.85 & 0.84/- & \textbf{0.92}/0.86 & \cellcolor{pink!20}0.89/\textbf{0.91} \\
		Cap        & 0.60/0.63 & 0.70/0.73 & 0.73/- & 0.74/0.76 & 0.80/0.93 & 0.75/0.90 & 0.77/- & 0.87/0.94 & \cellcolor{pink!20}\textbf{0.88}/\textbf{0.91} \\
		Cup        & 0.55/0.64 & 0.70/0.67 & 0.77/- & 0.82/0.77 & 0.85/0.92 & 0.91/0.92 & 0.85/- & \textbf{0.96}/0.92 & \cellcolor{pink!20}0.94/\textbf{0.95} \\
		Eraser     & 0.63/0.71 & 0.55/0.55 & 0.89/- & 0.90/0.71 & {0.99}/\textbf{0.97} & 0.60/0.95 & 0.77/- & \textbf{1.0}/0.95 & \cellcolor{pink!20}0.98/\textbf{0.97}\\
		Headset    & 0.60/0.58 & 0.70/0.55 & 0.75/- & 0.77/0.64 & 0.87/0.87 & 0.90/0.88 & 0.87/- & \textbf{0.94}/0.86 & \cellcolor{pink!20}0.92/\textbf{0.90} \\
		Helmet     & 0.49/0.58 & 0.60/0.63 & 0.70/- & 0.73/0.72 & 0.84/0.90 & 0.84/0.81 & 0.82/- & 0.93/0.89 & \cellcolor{pink!20}\textbf{0.94}/\textbf{0.92} \\
		Jar        & 0.44/0.54 & 0.78/0.77 & 0.84/- & 0.87/0.82 & 0.87/0.87 & \textbf{1.00}/0.96 & 0.97/- & 0.91/0.88 & \cellcolor{pink!20}0.98/\textbf{0.97} \\
		Micro.     & 0.36/0.36 & 0.76/0.74 & 0.76/- & 0.78/0.66 & 0.78/0.81 & \textbf{1.00}/\textbf{0.95} & 0.92/- & 0.92/0.83 & \cellcolor{pink!20}0.92/0.88 \\
		Shelf      & 0.56/0.55 & 0.60/0.61 & 0.70/- & 0.73/0.69 & 0.57/0.66 & 0.71/0.87 & \textbf{0.84}/- & 0.69/0.81 & \cellcolor{pink!20}0.78/\textbf{0.89} \\
		Tap        & 0.75/0.68 & 0.69/0.70 & 0.82/- & 0.77/0.54 & 0.71/0.74 & 0.84/0.89 & 0.96/- & 0.79/0.79 & \cellcolor{pink!20}\textbf{0.91}/\textbf{0.90} \\
		Vase       & 0.53/0.63 & 0.63/0.60 & 0.73/- & 0.73/0.69 & 0.83/0.94 & \textbf{0.93}/0.90 & 0.84/- & 0.89/0.94 & \cellcolor{pink!20}\textbf{0.93}/\textbf{0.97} \\
		\midrule
		Average    & 0.55/0.63 & 0.66/0.66 & 0.75/- & 0.78/0.69 & 0.84/0.88 & 0.85/0.89 & 0.89/- & 0.91/0.91 & \cellcolor{pink!20}\textbf{0.93}/\textbf{0.94} \\
		\bottomrule
	\end{tabular}
\end{table*}

\begin{table*}[t]
	\centering
	\footnotesize
	\caption{O-AUROC / P-AUROC scores for anomaly detection across 12 categories of Real3D-AD. Bold numbers indicate the best results.}
	\label{tab:real3d}
	\setlength{\tabcolsep}{2.8pt} 
	\renewcommand{\arraystretch}{1.1}
	\begin{tabular}{l|cccccccccc} 
		\toprule
		
		\textbf{Category} & \multicolumn{2}{c}{\textbf{PatchCore~\cite{roth2022towards}}} & \textbf{RegAD~\cite{huang2022registration}} & \textbf{ISMP~\cite{liang2025look}} & \textbf{PO3AD~\cite{ye2024po3ad}} & \textbf{Reg2Inv} & \textbf{MC3AD~\cite{cheng2025mc3d}} & \textbf{MC4AD~\cite{liang2025examining}} & \textbf{PASDF~\cite{Zheng2025Bridging3D}} & \multirow{2}{*}{\cellcolor{pink!20}\textbf{Ours}} \\
		
		\textbf{Source} & \multicolumn{2}{c}{\textbf{CVPR'22}} & \textbf{NeurIPS'23} & \textbf{AAAI'24} & \textbf{CVPR'25} & \textbf{NeurIPS'25} & \textbf{IJCAI'25} & \textbf{NeurIPS'25} & \textbf{ICCV'25} & \\
		
		\midrule
		\textbf{Feature} & FPFH & PointMAE  & PointMAE & Raw & Raw  & Raw & PointMAE & Raw & Raw & \cellcolor{pink!20!white}Raw \\
		\midrule
		Airplane  & \textbf{0.88}/0.56 & 0.73/0.57 & 0.72/0.63  & 0.86/0.75 & 0.80/-- & 0.82/\textbf{0.92} & 0.85/0.63 & 0.87/0.79 & 0.63/0.78 & \cellcolor{pink!20!white}0.87/0.81 \\
		Car       & 0.59/0.75 & 0.50/0.61 & 0.70/0.72 & 0.73/0.84 & 0.65/-- & 0.76/\textbf{0.94} & 0.75/0.82 & 0.68/0.84 & \textbf{0.96}/0.80 & \cellcolor{pink!20!white}0.82/0.91 \\
		Candy     & 0.54/0.78 & 0.66/0.63 & 0.69/0.72  & 0.85/0.91 & 0.79/-- & \textbf{1.00}/0.97 & 0.83/0.55 & 0.81/0.84 & 0.79/0.55 & \cellcolor{pink!20!white}\textbf{1.00}/\textbf{0.98} \\
		Chicken   & 0.84/0.43 & 0.83/0.73 & 0.85/0.68  & 0.71/0.80 & 0.69/-- & \textbf{0.94}/\textbf{0.91} & 0.72/0.64 & 0.70/0.85 & 0.74/0.77 & \cellcolor{pink!20!white}0.79/0.85 \\
		Diamond   & 0.57/0.83 & 0.78/0.72 & 0.90/0.84  & 0.95/0.93 & 0.80/-- & \textbf{1.00}/0.98 & 0.96/0.94 & 0.84/0.84 & 0.89/0.70 & \cellcolor{pink!20!white}\textbf{1.00}/\textbf{0.99} \\
		Duck      & 0.55/0.26 & 0.49/0.53 & 0.58/0.50  & 0.71/0.88 & 0.82/-- & 0.75/\textbf{0.94} & 0.83/0.82 & 0.82/0.78 & 0.66/0.77 & \cellcolor{pink!20!white}\textbf{0.92}/0.89 \\
		Fish      & 0.68/0.83 & 0.63/0.72 & 0.92/0.83  & \textbf{0.95}/0.89 & 0.86/-- & 0.67/0.85 & 0.86/\textbf{0.93} & 0.89/\textbf{0.93} & \textbf{0.99}/0.84 & \cellcolor{pink!20!white}0.77/0.89 \\
		Gemstone  & 0.37/0.91 & 0.37/0.44 & 0.42/0.55  & 0.47/0.86 & 0.69/-- & 0.74/0.91 & 0.56/0.46 & 0.70/0.89 & 0.63/0.65 & \cellcolor{pink!20!white}\textbf{0.76}/\textbf{0.94} \\
		Seahorse  & 0.51/0.74 & 0.54/0.63 & 0.76/0.82  & 0.73/0.81 & 0.76/-- & 0.53/0.65 & 0.72/0.66 & 0.75/0.82 & \textbf{1.00}/\textbf{0.89} & \cellcolor{pink!20!white}0.79/\textbf{0.89} \\
		Shell     & 0.59/0.74 & 0.50/0.71 & 0.58/0.81  & 0.62/0.84 & 0.80/-- & 0.69/0.91 & 0.80/0.78 & 0.80/0.87 & 0.85/0.65 & \cellcolor{pink!20!white}\textbf{0.88}/\textbf{0.94} \\
		Starfish  & 0.44/0.61 & 0.52/0.58 & 0.51/0.62  & 0.66/0.64 & 0.76/-- & \textbf{0.84}/0.84 & 0.77/0.69 & 0.77/0.79 & 0.62/0.70 & \cellcolor{pink!20!white}0.79/\textbf{0.87} \\
		Toffees   & 0.57/0.75 & 0.59/0.58 & 0.83/0.76  & 0.84/0.90 & 0.77/-- & 0.63/0.74 & 0.74/\textbf{0.93} & 0.79/0.78 & \textbf{0.87}/0.84 & \cellcolor{pink!20!white}0.75/0.86 \\
		\midrule
		Average   & 0.59/0.68 & 0.59/0.62 & 0.70/0.71 & 0.77/0.84 & 0.77/-- & 0.78/\textbf{0.88} & 0.78/0.77 & 0.79/0.84 & 0.80/0.75 & \cellcolor{pink!20!white}\textbf{0.83}/\textbf{0.89} \\
		\bottomrule
	\end{tabular}
\end{table*}

\begin{figure*}[t]
	\centering
	
	\includegraphics[width=\linewidth, height=0.3\textheight]{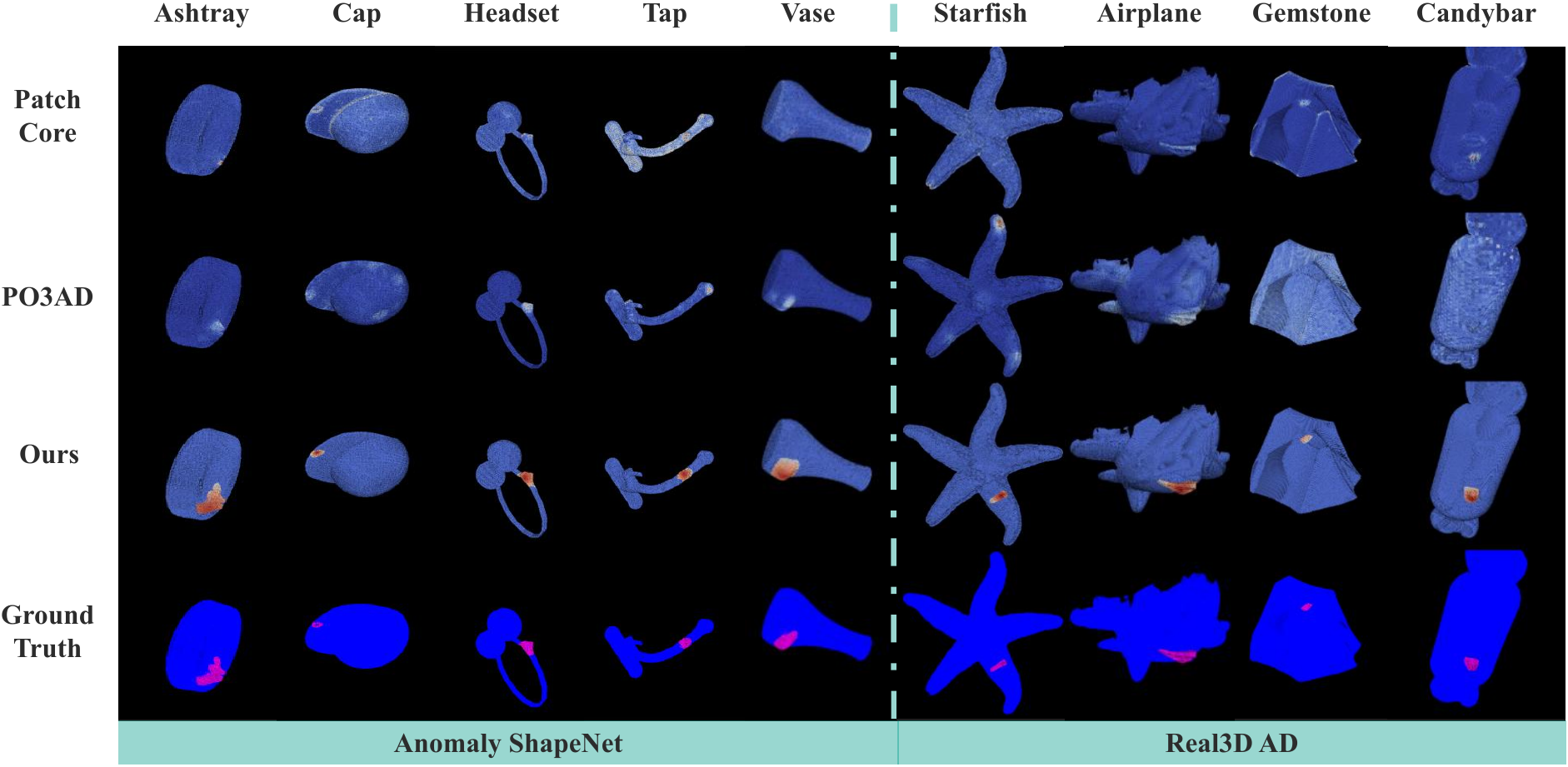} \vspace{-5mm}
	
	\caption{Visual comparison of detection results obtained by different methods on Anomaly-ShapeNet and Real3D-AD datasets. }
	\label{fig:compare2}
\end{figure*}

\subsection{Anomaly Point Cloud Generation}

\begin{figure*}[!h]
	\centering

	\includegraphics[width=\linewidth, height=0.4\textheight]{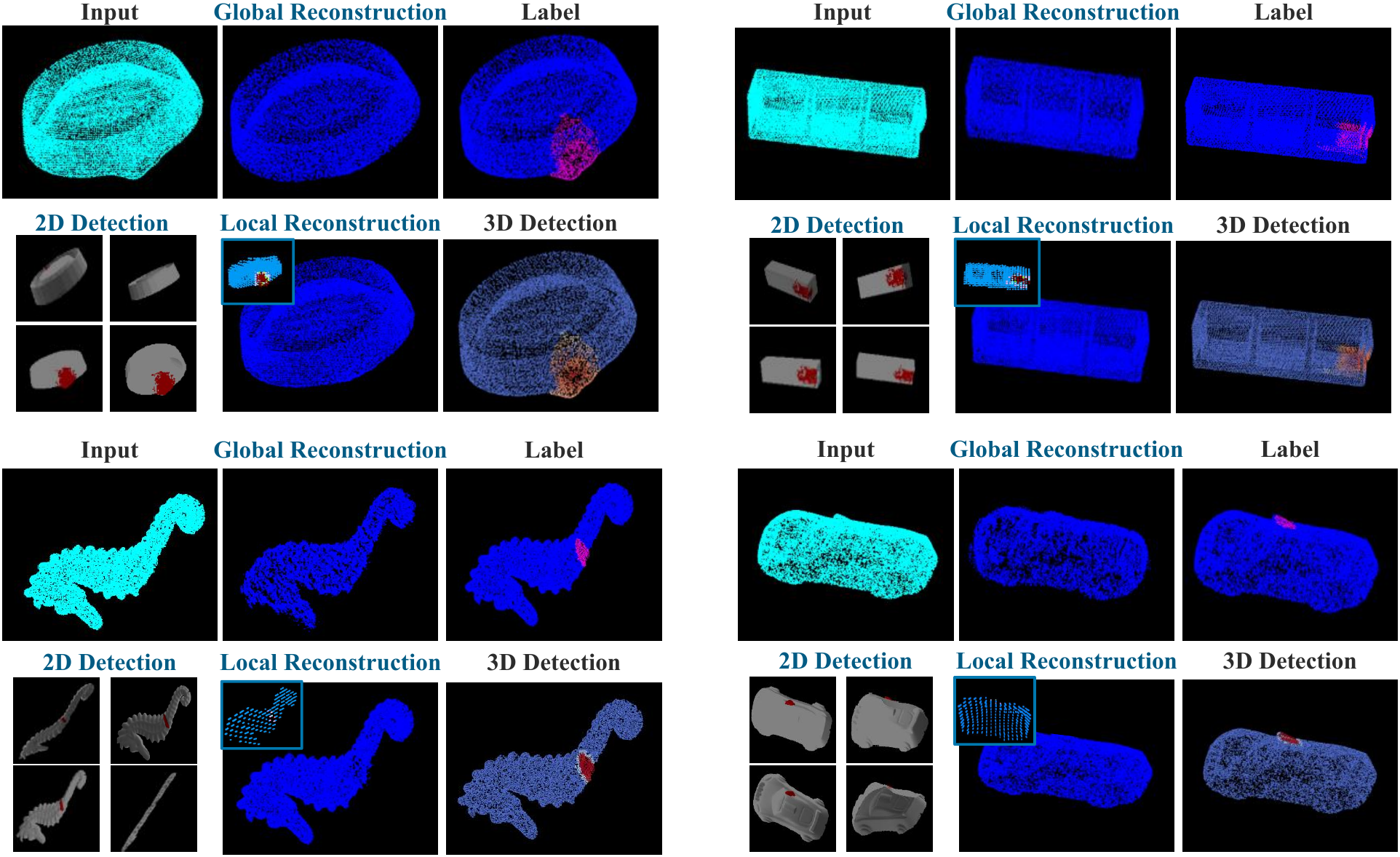}
	\vspace{-3mm}
	
	\caption{The visualization of reconstruction  and detection results on \textbf{Anomaly ShapeNet} and \textbf{Real3D-AD}  datasets. The top 2 are from Anomaly ShapeNet, and the bottom 2 are from Real3D-AD. The light blue voxel image in the local reconstruction shows the 2D back-projected 3D anomaly-guiding masks.}
	\label{fig:pipeline}
\end{figure*}

\textbf{Implementation Details}
For dataset augmentation, we construct a pool of all normal meshes and point clouds, applying structural synthesis from MC4AD and our module for subtle textures. Augmented samples are rendered, cropped, and extracted to generate 12 variants per sample, totaling 19,200 samples for Anomaly-ShapeNet and 7,812 for Real3D-AD. The data is split into training, validation, and test sets (4:1:1 ratio), with the latter used for generation evaluation.

For multi-view image generation, we extend image diffusion to a multi-view framework by horizontally concatenating frames to model geometric relationships. We employ instance attention (max 3 instances) after cross-attention. The 9-channel input comprises the noisy latent, mask, and masked reference image. Training uses the Adam optimizer ($\text{lr} = 1 \times 10^{-4}$) with a batch size of 16. For point cloud generation, our transformer-based architecture integrates instance attention after each multi-head attention block. We utilize a diffusion framework with `v prediction` and optimize via Adam ($\text{lr} = 1 \times 10^{-4}$, batch size 16).

\noindent\textbf{Baselines and Metrics.} We compare PCDiff against state-of-the-art anomaly synthesis methods: CutPaste~\cite{li2021cutpaste}, which performs point cloud injection; R3D-AD~\cite{zhou2024r3d}, which synthesizes structural defects (bumps, dents); and Gaussian Splatting (Gau-SP)~\cite{kerbl20233d}. For fair comparison, Gau-SP is conditioned on multi-view images generated by our pipeline. We evaluate 100 samples per category using two primary metrics: (1) CLIP Similarity, computed by averaging scores across 18 rendered views of the reconstructed mesh (via MeshAnything~\cite{chen2024meshanything}) against textual prompts, and (2) F-score, to assess the geometric fidelity between generated and ground-truth anomaly point clouds.

\noindent\textbf{Results} From Table~\ref{tab:generation}, PCDiff surpasses all baseline methods regarding F-Score and CLIP similarity. Additionally, we show the visual results on the Anomaly-ShapeNet dataset in Figure~\ref{fig:show_result}.  Our model demonstrates superior authenticity and fine-texture in anomaly generation. For example, in the `Bulge' category, anomalies exhibit convex geometry with subtle bumps, guided by an irregular texture map. In the `Concavity' category, defects resemble realistic concavities with rich sink texture patterns.  Especially, In the `Cap' category, the generated subtle scratch on mesh matches the guiding line texture map. Overall, PCDiff synthesis sophisticated textures and structures, reflecting miniature details such as scratches in both point clouds and meshes.


\subsection{Anomaly Detection}

\textbf{Implementation Details}
\noindent
For local reconstruction, the prior 2D anomaly detection backbone shares the same architecture as our generation backbone, without instance-level attention. The 3D point cloud reconstruction backbone is aligned with the generation backbone, with the instance-level attention enabled with 2D detected mask as condition.
For global reconstruction, the 3D point cloud backbone remains identical to the generation backbone, with instance attention disabled to reflect the absence of localized prompt cues.
For merged detection, the 3D point cloud backbone again mirrors the generation architecture with instance attention omitted.

\noindent\textbf{Baselines and Metrics} To validate the efficacy of our method, we evaluate it with state-of-the-art anomaly-detection methods, including BTF~\cite{horwitz2023back} (CVPR'23), M3DM~\cite{wang2023multimodal} (CVPR'23), PatchCore~\cite{roth2022towards} (CVPR'22), RegAD~\cite{huang2022registration} (NeurIPS'23), IMRNet~\cite{li2024towards} (CVPR'24), R3D-AD~\cite{zhou2024r3d} (ECCV'24), ISMP~\cite{liang2025look} (AAAI'24), POA3D~\cite{ye2024po3ad} (CVPR'25), PASDF~\cite{Zheng2025Bridging3D} (ICCV'25), Reg2Inv~\cite{yu2025registration} (NeurIPS'25), MC3AD~\cite{cheng2025mc3d}  and MC4AD~\cite{liang2025examining}(NeurIPS'25). Similar to ~\cite{ye2024po3ad}, we employ the O-AUROC  and P-AUROC metrics for anomaly-detection evaluation, where higher values reflect better detection performance.

\noindent\textbf{Results} The comparison results for the Anomaly-ShapeNet dataset are listed in Table~\ref{tab:anoshape}. The proposed model achieves a superior detection performance in anomaly detection, boasting the highest O-AUROC of $0.93$. 
Further validation is performed on the Real3D-AD dataset. Table~\ref{tab:real3d} shows that our method performs exceptionally well in anomaly localization with the highest O-AUROC of $0.82$. This validates the significant positive impact of our generated data on downstream anomaly inspection tasks.

Furthermore, we present a visual comparison of detection results on the Anomaly-ShapeNet and Real3D-AD datasets in Figure~\ref{fig:compare2}. As shown, PatchCore tends to highlight anomaly boundaries with relatively low anomaly scores and occasionally misses true anomalies, leading to false negatives (e.g., the vase) as well as false positives (e.g., the tap). PO3AD demonstrates improved coverage of anomaly regions but still produces false positive detections, particularly on challenging cases such as the  gemstone. In contrast, our method achieves more precise and consistent detection, capturing the majority of anomaly regions with high confidence while significantly reducing false positives. To provide deeper insight into the working mechanism, we further include representative intermediate results from our pipeline---namely global reconstruction, local reconstruction, and multi-view detection---on both Anomaly-ShapeNet and Real3D-AD, as illustrated in 
Figure~\ref{fig:pipeline}.

\subsection{Ablation Studies}

\noindent\textbf{Ablation Study on Generation.} We evaluate our generation framework using two complementary metrics: F-score for 3D point cloud fidelity and CLIP score for text-image alignment. Our ablation focuses on two core components: (1) transformation-based texture representation, which encodes local anomalies, and (2) multi-modal instance attention, which integrates textual, visual, and gradient signals.

For texture representation, we compare our proposed representation against traditional surface curvature~\cite{pauly2002efficient} across three variants. As shown in Table~\ref{tab:ablation_texture}, using Curvature alone yields an F-score of 0.83. Our \textbf{Gradient} representation alone achieves the highest F-score of 0.95, while the combination results in 0.91. These results underscore the superiority of our gradient module in capturing complex anomaly textures.

\begin{table}[!t]
	\centering
	\caption{Ablation study of texture representation on Anomaly ShapeNet datasets.}
	\vspace{-2mm}
	\setlength{\tabcolsep}{4pt}
	\begin{tabular}{c|c|c|c}
		\toprule
		& Curvature & Gradient & Curv. + Grad. \\
		\midrule
		F-score / CLIP & 0.83 / 0.82  & \textbf{0.95} / \textbf{0.81}  & 0.91 / 0.78 \\
		
		\bottomrule
	\end{tabular}
	\label{tab:ablation_texture}
	\vspace{-3mm}
\end{table}

\begin{table}[h]
	\centering
	\caption{Ablation study of the proposed PCDiff.}
	\vspace{-2mm}
	\setlength{\tabcolsep}{4pt}
	\begin{tabular}{c|c|cc|c|c}
		\toprule
		& Global Recon. & \multicolumn{2}{c|}{Local Recon.} & {Ano. Mask} & O-AUROC \\
		& & elevation & azimuth num & & \\
		\midrule
		a & $\checkmark$ & $\times$ & $\times$ & $\times$ & 0.84 \\
		b & $\checkmark$ & $[-45,0,45]$ & 6 & $\times$ & 0.90 \\
		c & $\checkmark$ & $[-45,0,45]$ & 6 & $\checkmark$ & \textbf{0.93} \\
		\midrule
		d & $\checkmark$ & $[-60,0,60]$ & 6 & $\checkmark$ & 0.87 \\
		e & $\checkmark$ & $[-30,0,30]$ & 6 & $\checkmark$ & 0.86 \\
		f & $\checkmark$ & $[-45,0,45]$ & 8 & $\checkmark$ & 0.90 \\
		g & $\checkmark$ & $[-45,0,45]$ & 4 & $\checkmark$ & 0.88 \\
		\midrule
		h & $\times$ & $[-45,0,45]$ & 6 & $\checkmark$ & 0.87 \\ 
		\bottomrule
	\end{tabular}
	\label{tab:ablation_local_global}
	\vspace{0.2cm}
\end{table}

\noindent\textbf{Ablation on Detection.} We evaluate our local-global detection framework on the Anomaly-ShapeNet dataset to quantify the contribution of each component. The baseline utilizes a standard global 3D anomaly reconstruction pipeline. We incrementally integrate the local reconstruction branch and 2D-based anomaly masks. Furthermore, we investigate the impact of image rendering hyperparameters---specifically elevation (angle relative to the horizontal plane) and azimuth (circular viewpoints)---on detection performance.

As detailed in Table~\ref{tab:ablation_local_global}, the global baseline achieves an O-AUROC of 0.84. Integrating the local reconstruction branch (b) improves performance by $0.06$. The inclusion of the anomaly mask provides an additional $0.03$ gain. Sensitivity analysis of viewpoint hyperparameters reveals that narrowing or broadening the elevation range (e.g., to $\pm 60^\circ$ or $\pm 30^\circ$) reduces O-AUROC by $0.03$--$0.04$. Similarly, deviating from the optimal azimuth count degrades results, likely due to redundant view overlap or insufficient coverage. Notably, removing the global reconstruction module (row h) causes a significant drop ($0.93 \to 0.78$), confirming that the global branch provides the essential structural anchors required for consistent local refinement.

\begin{figure}[!h]

	\includegraphics[width=\linewidth, height=0.3\textheight]{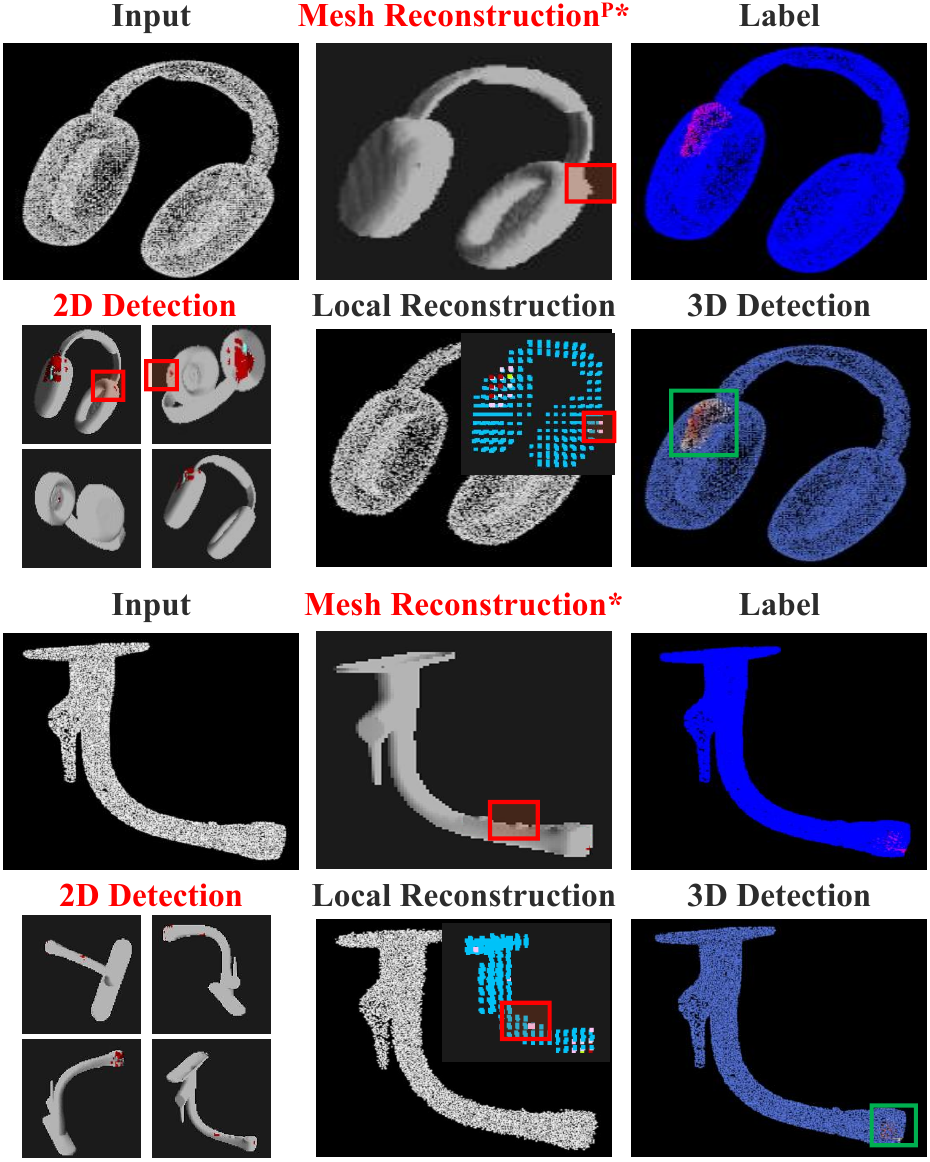}
	\vspace{-3mm}
	
	\caption{The visualization of failure mesh reconstruction and 2D detection results on Anomaly ShapeNet datasets, where the false positive accumulate to the voxel anomaly mask, however correctly reconstructed with point local reconstruction and yields accurate detection result.}
	\label{fig:failure}
\end{figure}

\noindent\textbf{Robustness to Unreliable Mesh Reconstruction and 2D Detection.} We investigate how inaccurate mesh reconstructions and imprecise 2D detection masks impact final performance. As shown in Fig.~\ref{fig:failure}, although errors can propagate, their influence is largely confined to the mask guidance stage. To mitigate error accumulation, we employ a dual strategy: (1) \textit{Strategic Data Augmentation}, intentionally incorporating normal regions into guidance masks during training to enhance robustness to imprecise 2D priors, and (2) \textit{Local Point Reconstruction}, which ``smooths'' geometric inconsistencies from the initial coarse mesh. Quantitative results confirm that our system maintains high accuracy even with suboptimal early-stage outputs, effectively decoupling the final detection from the failure modes of traditional end-to-end 3D reconstruction.

\noindent\textbf{Computational Efficiency and Deployment.} For industrial implementation, real-time detection latency is the critical determinant for edge deployment. Our global and local reconstruction Transformer, optimized with \textit{xFormers} and gradient checkpointing, achieves a reconstruction latency of only \textbf{2.92\,s}—a 26.6$\%$ improvement over MC3AD's 3.98\,s—despite the increased architectural depth required for 3D anomaly detection. While our model utilizes higher peak memory (17,312\,MB vs.\ 4,437\,MB), it delivers a significantly more efficient computational profile, requiring only \textbf{4,936\,GFLOPs} compared to the 17,271\,GFLOPs demanded by R3D-AD.

The auxiliary pipeline—comprising mesh reconstruction (0.62\,s), multi-view rendering (0.15\,s), 2D detection (0.57\,s), and back-projection (0.01\,s)—adds a marginal 1.35\,s, which is further mitigable via multi-process parallelism. Ultimately, our framework offers a favorable accuracy--efficiency trade-off: delivering a substantial performance gain (\textbf{0.82 vs.\ 0.79 AUROC}) while simultaneously reducing overall inference latency.

\section{Conclusion}

\label{sec:conclusion}

This paper proposed PCDiff, a gradient-guided point cloud diffusion method, for 3D anomaly generation and detection. It incorporates four key innovations: a high-fidelity anomaly generation, a gradient-based texture representation, a multi-modal instance attention mechanism, and a local-global joint detection framework. In 3D anomaly generation, the gradient-based texture representation improves the realism and diversity of weak anomalies, while the multi-modal instance attention mechanism enhances controllability over anomaly location, category, and texture. In 3D anomaly detection, the local-global joint detection framework enables fine-grained local anomaly reconstruction guided by 2D priors, while ensuring unbiased global geometric estimation. Extensive experiments validated the effectiveness of PCDiff, demonstrating significant improvements over state-of-the-art methods. 

\small
\bibliographystyle{IEEEtran}
\bibliography{egbib.bib}

\newpage

\vfill

\end{document}